\documentclass[letterpaper, 10 pt, conference]{ieeeconf}  

\IEEEoverridecommandlockouts                              

\usepackage{hyperref}

\usepackage{svg}
\usepackage{times}
\usepackage{helvet}
\usepackage{helvet}
\usepackage{tikz}
\usepackage{courier}
\usepackage{url}
\usepackage{graphicx}
\usepackage{cite}
\usepackage{caption}

\usepackage{amsmath}
\usepackage{algorithm}
\usepackage{algpseudocode}

\usepackage{graphics} 
\usepackage{epsfig} 
\usepackage{mathptmx} 
\usepackage{times} 
\usepackage{xspace}
\usepackage{calc} 
\usepackage{amsmath} 
\usepackage{amssymb}
\usepackage[capitalize]{cleveref}
\usepackage{xspace}
\let\labelindent\relax
\usepackage{enumitem}
\usepackage{siunitx}
\usepackage[utf8]{inputenc}
\usepackage{tabularx}
\usepackage{booktabs}
\usepackage{svg}
\usepackage{arydshln}

\usepackage[utf8]{inputenc}
\usepackage[T1]{fontenc}
\usepackage{textcomp}

\usepackage{booktabs}
\usepackage{multirow}
\usepackage[utf8]{inputenc}
\usepackage{graphicx}
\usepackage{caption}
\usepackage{subcaption}

\usepackage{xcolor}

\title{\LARGE \bf
Exploring Vision-Language Models for Open-Vocabulary Zero-Shot  Action Segmentation}

\author{Asim Unmesh$^{1}$, Kaki Ramesh$^{2}$, Mayank Patel$^{1}$, Rahul Jain$^{1}$, and Karthik Ramani$^{1}$%
\thanks{$^{1}$Purdue University, USA}%
\thanks{$^{2}$Birla Institute of Technology and Science (BITS) Hyderabad, India}%
}
\usepackage{xcolor} 
\usepackage{pifont}

\newcommand{\cmark}{{\color{blue}\ding{51}}} 

\newcommand{\xmark}{\ding{55}}

\begin{document}

\maketitle
\thispagestyle{empty}
\pagestyle{empty}


\begin{abstract}
Temporal Action Segmentation (TAS) requires dividing videos into action segments, yet the vast space of activities and alternative breakdowns makes collecting comprehensive datasets infeasible. Existing methods remain limited to closed vocabularies and fixed label sets. In this work, we explore the largely unexplored problem of \textbf{Open-Vocabulary Zero-Shot Temporal Action Segmentation (OVTAS)} by leveraging the strong zero-shot capabilities of Vision–Language Models (VLMs). We introduce a training-free pipeline that follows a segmentation-by-classification design: (i) \textit{Frame–Action Embedding Similarity (FAES)} matches video frames to candidate action labels, and (ii) \textit{Similarity-Matrix Temporal Segmentation (SMTS)} enforces temporal consistency. Beyond proposing OVTAS, we present a systematic study across 14 diverse VLMs, providing the first broad analysis of their suitability for open-vocabulary action segmentation. Experiments on standard benchmarks show that OVTAS achieves strong results without task-specific supervision, underscoring the potential of VLMs for structured temporal understanding. To foster further research, we release code and embeddings at \href{https://aunmesh.github.io/ovtas/}{our project page}.
\end{abstract}

\section{Introduction}



\begin{figure}[t]
    \centering
    \begin{subfigure}{\linewidth}
        \centering
        \includegraphics[width=\linewidth]{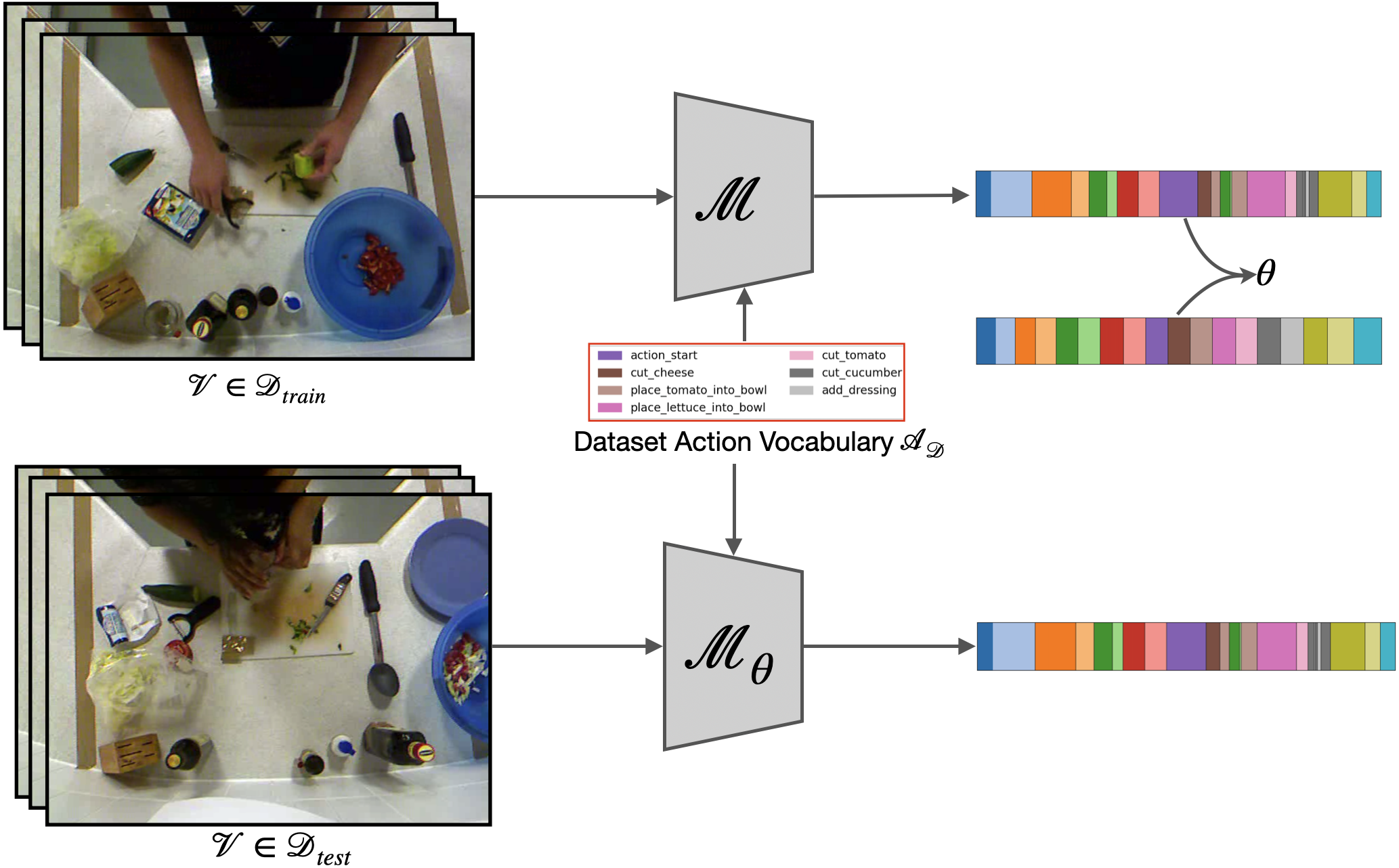}
        \caption{Other approaches}
        \label{fig:comparison_a}
    \end{subfigure}
    
    \vspace{3mm}
    
    \begin{subfigure}{\linewidth}
        \centering
        \includegraphics[width=\linewidth]{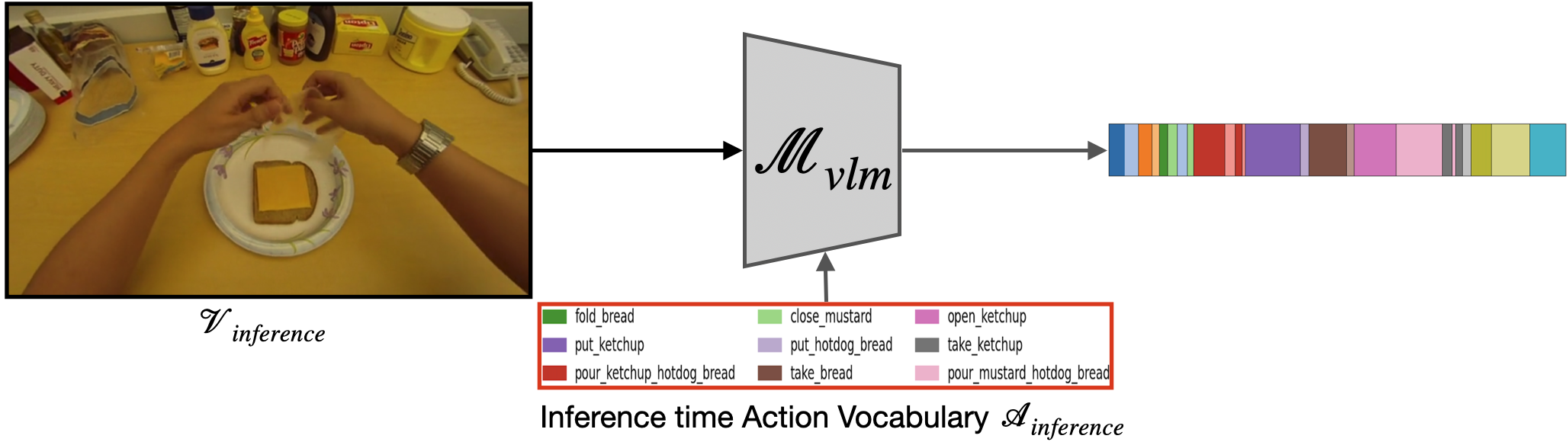}
        \caption{Proposed approach}
        \label{fig:comparison_b}
    \end{subfigure}
    
    \caption{ \textbf{Problem Setup:} Existing approaches in ~\ref{fig:comparison_a} are fixed vocabulary and do not generalize to unseen videos. 
    ~\ref{fig:comparison_b} illustrates our proposed method for open-vocabulary and zero-shot action segmentation.}
    \label{fig:comparison}
\end{figure}

Temporal Action Segmentation (TAS) has been an active area of research \cite{ding2023temporal} with applications in human activity understanding \cite{romeo2025multi}, surgical robotics \cite{de2021first}, robot task learning \cite{chen2025backbone}, and action assessment \cite{okamoto2024hierarchical}. The goal in action segmentation is to assign action labels to every frame of a video, segmenting it into meaningful units. Despite substantial progress, existing TAS methods remain constrained to a \textit{closed vocabulary} of action labels. Models are typically trained and evaluated on the same fixed set of classes, limiting their ability to generalize to new actions or unseen domains.

This closed-vocabulary assumption is particularly restrictive because the space of possible action vocabularies is vast: a single activity may be decomposed into dozens of steps, and each domain---such as kitchen tasks, assembly, or surgery---contains hundreds of distinct activities. Even for the same task, alternative segmentations may exist, reflecting different emphases such as object-centered versus process-centered perspectives. Constructing annotated datasets that cover this variability is infeasible, and as a result, closed-vocabulary TAS methods struggle to scale beyond label spaces defined in the dataset.

Inspired by the application of Vision–Language Models (VLMs) \cite{ju2022prompting, yu2025building} for action understanding tasks, we propose the \textbf{Open-Vocabulary Temporal Action Segmentation (OVTAS) pipeline}, a training-free, zero-shot approach that leverages the open-vocabulary and zero-shot capabilities of VLMs. Contrastively trained VLMs such as CLIP and SigLIP demonstrate strong image-level zero-shot recognition of novel categories by aligning visual and textual embeddings. Our key idea is to use this capability for TAS in a ``segmentation-by-classification'' setting.

However, VLM predictions at the frame level are temporally inconsistent, as they operate independently across frames. To overcome this, OVTAS follows a two-stage segmentation-by-classification design. Stage 1, \textit{Frame--Action Embedding Similarity (FAES)}, computes similarities between frame embeddings and text embeddings of action labels. Stage 2, \textit{Similarity-Matrix Temporal Segmentation (SMTS)} \cite{xu2024temporally}, decodes these similarities into temporally consistent label sequences using an optimal transport-based approach. 

Understanding how different Vision–Language Models behave in the task of open-vocabulary temporal action segmentation, informs future research and application about the choices of VLM models and their sizes. Given the various VLM models, varying across families (such as CLIP, SigLIP) and model sizes, a systematic exploration is needed to inform the research community of performance levels across different choices. To address this, we systematically explore 14 diverse VLMs across families and sizes, uncovering performance trends.


Our contributions are as follows:
\begin{itemize}
\item \textbf{Pipeline Design}: We introduce the \textbf{OVTAS pipeline}, a two-stage framework---FAES followed by SMTS---that produces temporally consistent per-frame labels without any task-specific training or fine-tuning. 
\item \textbf{Comprehensive VLM Study}: We extensively evaluate state-of-the-art VLM families across three TAS benchmarks, uncovering performance trends and key factors influencing success in open-vocabulary segmentation.
\end{itemize}

To enable further research, we release our codebase and the extracted vision–language embeddings of 14 VLMs for all three datasets. Since feature extraction from large VLMs demands substantial computational resources, providing ready-to-use embeddings eases this barrier, enabling broader research into VLMs for Action Segmentation and other action understanding tasks.

\section{Related Works}

\begin{table*}[htbp]
\centering
\renewcommand{\arraystretch}{1.2}
\setlength{\tabcolsep}{4pt}
\caption{Comparison of our method with other methods in TAS based on generalization ability and training data needs. \cmark denotes applicability.}
\label{tab:tas_supervision}
\begin{tabular}{l|c|c|c|c|c|c}
\toprule
Method & Zero-Shot & Open-Vocab & Unsupervised & Weakly Sup. & Semi-Sup. & Training-Free \\
\midrule
ASAL \cite{li2021action}     & \xmark & \xmark & \cmark & \xmark & \xmark & \xmark \\
COIN-SSL \cite{ding2022leveraging}    & \xmark & \xmark & \xmark & \xmark & \cmark & \xmark \\
NN-Viterbi \cite{richard2018neuralnetwork}   & \xmark & \xmark & \xmark & \cmark & \xmark & \xmark \\
ASAL-SSL \cite{li2021action}         & \xmark & \xmark & \xmark & \xmark & \cmark & \xmark \\
TCCNet \cite{khan2022timestamp}       & \xmark & \xmark & \xmark & \cmark & \xmark & \xmark \\
UDE \cite{swetha2021unsupervised}        & \xmark & \xmark & \cmark & \xmark & \xmark & \xmark \\
CTC \cite{richard2017weakly}          & \xmark & \xmark & \xmark & \cmark & \xmark & \xmark \\
D3TW \cite{li2020set}                 & \xmark & \xmark & \xmark & \cmark & \xmark & \xmark \\
ASRF-unsup. \cite{ishikawa2021alleviating} & \xmark & \xmark & \cmark & \xmark & \xmark & \xmark \\
C2F-TCN \cite{xu2024temporally}       & \xmark & \xmark & \cmark & \xmark & \xmark & \xmark \\
HTK \cite{kuehne2017weakly}           & \xmark & \xmark & \xmark & \cmark & \xmark & \xmark \\
BCN \cite{wang2020boundary}           & \xmark & \xmark & \xmark & \cmark & \xmark & \xmark \\
HVQ \cite{spurio2025hierarchical}      & \xmark & \xmark & \cmark & \xmark & \xmark & \xmark \\
U-OT \cite{xu2024temporally}      & \xmark & \xmark & \cmark & \xmark & \xmark & \xmark \\
\midrule
\textbf{OVTAS (Ours)}                 & \cmark & \cmark & \xmark & \cmark & \xmark & \cmark \\
\bottomrule
\end{tabular}
\end{table*}

\label{related_works}
Temporal Action Segmentation (TAS) has been studied under different supervision regimes.
\textbf{Weakly Supervised Methods.} These approaches aim to reduce the need for dense frame-level annotation by exploiting weaker forms of supervision that are cheaper to obtain. Examples include (a) timestamp supervision \cite{li2021temporal, khan2022timestamp}, where only a few time instances are annotated; (b) action set supervision \cite{richard2018action, li2020set}, where only the unordered list of actions is provided; and (c) transcript supervision \cite{richard2017weakly, kuehne2017weakly}, where the sequence of actions is known but not aligned to frames. Among these, timestamp supervision typically yields the best performance, followed by action sets, and then transcripts.
\textbf{Semi-Supervised Methods.} Semi-supervised methods \cite{ding2022leveraging, gammulle2020fine, ding2022leveraging} use dense labels for only a subset of videos, showing that even a small number of fully labeled videos can outperform weak supervision on larger datasets.
\textbf{Unsupervised Methods.} Unsupervised TAS methods \cite{xu2024temporally, bueno2023leveraging, wang2022sscap, song2025unsupervised} operate without any action-level labels, relying only on video-level activity labels.
\textbf{Our Approach.} In contrast, our OVTAS pipeline explores a new regime: \textit{training-free, zero-shot, open-vocabulary TAS}. By leveraging VLMs, OVTAS segments videos without any task-specific training and generalizes to unseen action labels, addressing the closed-vocabulary constraint in prior work. A comparison of existing temporal action segmentation approaches is given in \cref{tab:tas_supervision}.

\section{Method}

\begin{figure}[t]
  \centering
  \includegraphics[width=0.8\linewidth]{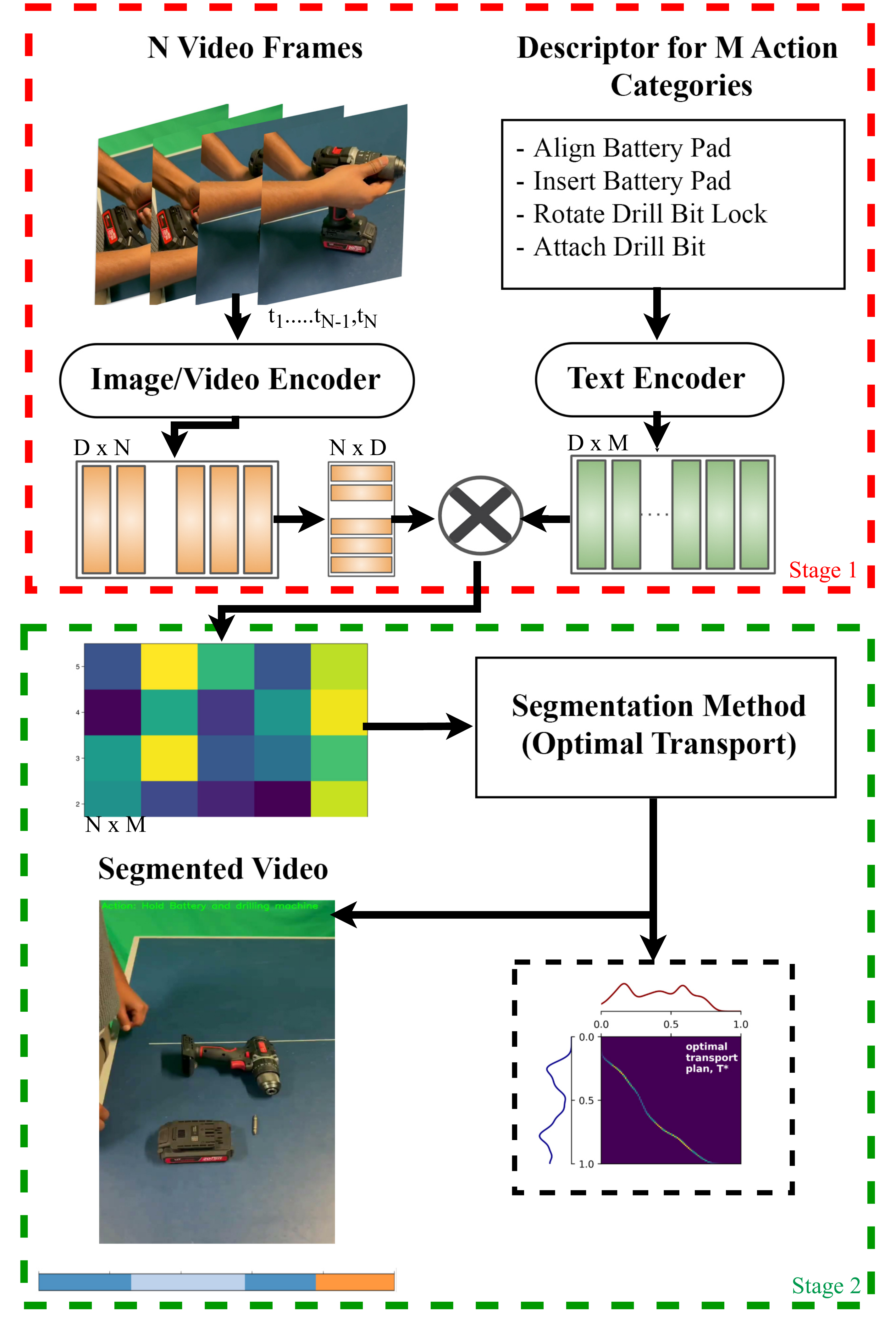}
  \caption{Open-Vocabulary Temporal Action Segmentation (OVTAS) Pipeline. 
  Our 2-stage pipeline adopts a ``segmentation by classification'' approach to tackle temporal action segmentation (TAS). 
  Stage 1, Frame–Action Embedding Similarity (FAES), generates a similarity matrix by matching frames with action labels. 
  Stage 2, Similarity-Matrix driven Temporal Segmentation (SMTS), uses optimal transport with a temporal prior to enforce temporal consistency, producing stable action segments.}
  \label{fig:tpgas_pipeline}
\end{figure}

\subsection{Overview}

We propose \textbf{Open-Vocabulary Temporal Action Segmentation (OVTAS)}, a \textit{training-free, zero-shot} pipeline that requires only a set of candidate action labels (the \emph{action set supervision}) and video frames as inputs. 
Action set supervision assumes that, given knowledge of the high-level task or activity (e.g., ``making tea''), we have access to the set of all possible fine-grained actions (e.g., ``boil water,'' ``pour tea,'' ``add sugar'') but not their order or boundaries. 
Our method proceeds in two stages: (i) \textit{Frame–Action Embedding Similarity (FAES)} computes similarities between video frames and action label embeddings, and (ii) \textit{Similarity-Matrix driven Temporal Segmentation (SMTS)} decodes these similarities into temporally consistent action segments using entropy-regularized optimal transport with a temporal prior.

\subsection{Terminology}

\noindent\textbf{Notation.} 
Bold uppercase = matrices, bold lowercase = vectors, italics = scalars. 
Let $T$ be the number of frames, $N$ the number of action labels, and $C$ the embedding dimension.

\begin{description}
  \item[A (Activity)] 
  A high-level category for an entire video (e.g., ``making tea'').

  \item[$\mathbf{X}\in\mathbb{R}^{T\times C}$ (Frame Embeddings)] 
  Row-wise $\ell_2$-normalized frame embeddings from the VLM vision encoder; 
  $\|\mathbf{x}_t\|_2=1$.

  \item[$\mathbf{A}\in\mathbb{R}^{N\times C}$ (Action Embeddings)] 
  Row-wise $\ell_2$-normalized text embeddings for the $N$ action labels from the VLM text encoder; 
  $\|\mathbf{a}_n\|_2=1$.

  \item[$\mathbf{S}\in\mathbb{R}^{T\times N}$ (Similarity Matrix)] 
  Cosine similarity (dot product) between frame and action embeddings:
  \[
    \mathbf{S} = \mathbf{X}\,\mathbf{A}^\top, 
    \quad \ell_t = \mathbf{x}_t\,\mathbf{A}^\top \in \mathbb{R}^{1\times N}.
  \]

  \item[$\mathbf{P}\in\mathbb{R}^{T\times N}$ (Frame Classification Probabilities)] 
  Row-wise softmax over actions:
  \[
    \mathbf{P} = \mathrm{softmax}_N(\mathbf{S}), 
    \quad \mathbf{p}_t = \mathrm{softmax}(\ell_t).
  \]
\end{description}

\subsection{Preliminaries: Optimal Transport Decoder}

We adopt the Action Segmentation OT (ASOT) decoder of Xu and Gould~\cite{xu2024temporally}. 
Given frame–action similarities $\mathbf{S}\in\mathbb{R}^{T\times N}$, we define a visual cost $\mathbf{C}=\mathbf{1}-\mathbf{S}$ and a diagonal temporal prior $R$ with
\[
R_{ij}=\Bigl|\tfrac{i}{T}-\tfrac{j}{N}\Bigr|,
\]
which encourages monotone alignment. 
We then solve for a coupling $\Pi\in\mathbb{R}^{T\times N}$:
\begin{equation}
\label{eq:asot_brief}
\Pi^\star=\arg\min_{\Pi\in U(\mathbf{u},\mathbf{v})}
\;\langle \Pi,\;\mathbf{C}+\rho R\rangle 
\;-\;\varepsilon\,H(\Pi),
\quad 
\mathbf{u}=\tfrac{1}{T}\mathbf{1}_T,\;\mathbf{v}=\tfrac{1}{N}\mathbf{1}_N,
\end{equation}
where $U(\mathbf{u},\mathbf{v})$ is the transport polytope and $H(\Pi)$ the entropy of $\Pi$. 
We use the same regularization and solver style as \cite{xu2024temporally}, and decode frame labels by
\[
\hat{y}_t=\arg\max_j \Pi^\star_{t,j}.
\]

\textit{Instantiations for OVTAS.} 
(i) \textbf{Open-vocabulary costs:} $\mathbf{C}$ is derived from VLM similarities (FAES). 
(ii) \textbf{Action-set supervision:} only the set of actions is assumed known, and their order is randomized when constructing the temporal prior $R$. 
(iii) \textbf{Hyperparameters:} $(\varepsilon,\rho,\ldots)$ are chosen via grid search on a small set of selected videos per dataset; the selected values are fixed for all experiments.

\subsection{Stage 1: Frame–Action Embedding Similarity (FAES)}

FAES computes raw similarity scores between frame and action embeddings:
\begin{enumerate}
  \item \textbf{Prompt Construction.} 
  Each action label is normalized into a natural-language phrase (e.g., ``pour\_coffee'' $\rightarrow$ ``pour coffee''). 
  These normalized labels are tokenized and encoded by the VLM text encoder. 

  \item \textbf{Embed frames and actions.} 
  Obtain $\mathbf{X}\in\mathbb{R}^{T\times C}$ and $\mathbf{A}\in\mathbb{R}^{N\times C}$ from the VLM encoders, applying row-wise $\ell_2$-normalization.

  \item \textbf{Compute similarities.} 
  \[
    \mathbf{S} = \mathbf{X}\,\mathbf{A}^\top \in \mathbb{R}^{T\times N}, 
    \quad \ell_t=\mathbf{x}_t\mathbf{A}^\top.
  \]
\end{enumerate}

\subsection{Stage 2: Similarity Matrix–driven Temporal Segmentation (SMTS)}

SMTS consumes the similarity matrix $\mathbf{S}$ and produces temporally consistent segments:
\begin{enumerate}
  \item \textbf{Action Order (Action Set Supervision).} 
  We assume action set supervision, we know only the set of actions present in the video, not their order. While constructing the similarity matrix, the ordering of the action embeddings is random. Action-set supervision is a reasonable as knowing activity typically implies knowing the set of possible actions.

  \item \textbf{OT Alignment.} 
  We apply the ASOT decoder (Eq.~\ref{eq:asot_brief}) with visual cost $\mathbf{C}$ and temporal prior $R$, yielding a coupling $\Pi^\star$. 
  The entropy term ensures the problem is convex and yields a unique solution. 
  In practice, $\Pi^\star$ is computed via Sinkhorn \cite{knight2008sinkhorn} iterations with log-stabilization, which scales linearly in $T\!\times\!N$ per iteration.

  \item \textbf{Action Segmentation.} 
  Assign each frame $t$ the action with maximum OT mass:
  \[
    \hat y_t = \arg\max_{j}\,\Pi^*_{t,j}.
  \]
\end{enumerate}

\begin{table}[htbp]
\centering
\renewcommand{\arraystretch}{1.15}
\setlength{\tabcolsep}{5pt}
\caption{Video duration statistics in seconds.}
\label{tab:video_durations}

\begin{tabular}{l|c c c}
\toprule
\textbf{Dataset} & Min (s) & Max (s) & Mean (s) \\
\midrule
GTEA      & 42.27  & 133.93 & 74.35  \\
Breakfast & 12.27  & 649.53 & 137.40 \\
50 Salads & 251.60 & 605.00 & 385.25 \\
\bottomrule
\end{tabular}
\end{table}

\begin{table}[htbp]
\centering
\renewcommand{\arraystretch}{1.15}
\setlength{\tabcolsep}{5pt}
\caption{Number of ground-truth segments per video.}
\label{tab:segment_counts}
\begin{tabular}{l|c c c}
\toprule
\textbf{Dataset} & Min & Max & Mean \\
\midrule
GTEA      & 24 & 49 & 36.3 \\
Breakfast & 1  & 18 & 5.4  \\
50 Salads & 15 & 27 & 20.7 \\
\bottomrule
\end{tabular}
\end{table}

\section{Experiments}

\subsection{Setup}

\subsubsection{Datasets} 
Experiments are performed on three standard Action Segmentation Benchmarks: Breakfast, 50 Salads, and the Georgia Tech Egocentric Activities (GTEA) dataset~\cite{fathi2011learning,stein2013combining,kuehne2014language}. 
\textbf{Breakfast}~\cite{kuehne2014language} contains 1,712 videos (77 hours total) of 52 participants performing ten breakfast-related activities (e.g., making coffee, frying eggs). The recordings are captured from 3 to 5 third-person cameras, with annotations for 48 fine-grained actions. 
\textbf{50 Salads}~\cite{stein2013combining} includes 50 top-view videos (5.5 hours total) of subjects preparing mixed salads, annotated with 17 action classes such as cutting, mixing, and adding ingredients. 
\textbf{GTEA}~\cite{fathi2011learning} consists of 28 egocentric videos (0.4 hours total) of four participants preparing seven different meals, annotated with 71 action classes. The statistics for video lengths of each dataset is mentioned in \cref{tab:video_durations}, and number of action segments  in the annotated ground truth is mentioned in \cref{tab:segment_counts}.
We evaluate each dataset using its official evaluation splits and report the final results as the average across splits.

\subsubsection{Metrics} 
We use 5 key metrics to report the result: F1@10, F1@25, F1@50, Accuracy, and Edit Scores. 
While Accuracy measures the overall proportion of correctly labeled frames, the F1 scores at different overlap thresholds capture the quality of segment-wise alignment with the ground truth, balancing precision and recall. 
Edit Score evaluates the temporal ordering and continuity of predicted segments, penalizing over-segmentation and fragmentation. 
Together, these metrics provide a comprehensive evaluation of action segmentation performance.

\subsubsection{Baselines}
We construct four training–free, zero–shot, open–vocabulary baselines. Let a video of $T$ frames have frame embeddings $F=\{f_t\}_{t=1}^T\in\mathbb{R}^{T\times D}$ and a label set $\mathcal{C}$ of size $C$ with text (or label) embeddings $A=\{a_c\}_{c\in\mathcal{C}}\in\mathbb{R}^{C\times D}$. We use cosine similarity
\[
S_{t,c}\;=\;\Big\langle \tfrac{f_t}{\|f_t\|_2},\; \tfrac{a_c}{\|a_c\|_2}\Big\rangle \in[-1,1],
\]
and report per–frame predictions $y_{1:T}$.

\paragraph{(1) Random–Uniform (RU).}
Each frame is labeled by a uniform categorical draw over classes:
\[
y_t \sim \mathrm{Cat}\!\left(\pi\right),\qquad \pi_c=\tfrac{1}{C}\ \ \forall c\in\mathcal{C},\ \ t=1,\dots,T.
\]
(If a background class exists, it is included in $\mathcal{C}$ as usual.)

\paragraph{Equal–Splits family.}
We take inspiration from \cite{sarfraz2021temporally}, and construct 3 equal splits baselines. For constructing an equal splits baseline, we partition the timeline into $K$ contiguous bins with edges
\[
e_k=\Big\lfloor \tfrac{kT}{K}\Big\rfloor,\quad k=0,\dots,K,\qquad
B_k=\{e_{k}+1,\dots,e_{k+1}\}.
\]
A single label $\hat y_k$ is chosen per bin and then expanded to frames: $y_t=\hat y_k$ for all $t\in B_k$. We use three training–free labeling rules:

\paragraph{(2) Equal–Splits Mean (ES–Mean).}
Score each class by its mean similarity in the bin and take the argmax:
\[
s^{\text{mean}}_{k,c}\;=\;\frac{1}{|B_k|}\sum_{t\in B_k} S_{t,c},\qquad
\hat y_k \;=\; \arg\max_{c\in\mathcal{C}} \ s^{\text{mean}}_{k,c}.
\]

\paragraph{(3) Equal–Splits Vote (ES–Vote).}
First take per–frame winners $\tilde y_t=\arg\max_{c} S_{t,c}$, then choose the modal class in the bin:
\[
\hat y_k \;=\; \operatorname*{mode}\big\{\tilde y_t:\ t\in B_k\big\}.
\]
(Ties, if any, are broken by larger $s^{\text{mean}}_{k,c}$.)

\paragraph{(4) Equal–Splits Non–Repetition Penalty (ES–NRP).}

Equal–splits can collapse to a trivial solution where every bin receives the same label. The non–repetition penalty adds a small cost for assigning the same label to adjacent bins, discouraging this degenerate constant prediction.

Let $s_{k,c}$ be any fixed per–bin class score (we use $s^{\text{mean}}_{k,c}$ above). We decode bin labels by a simple dynamic program that discourages repeating the same class in adjacent bins:
\[
\hat y_{1:K}
\;=\;
\arg\max_{y_1,\dots,y_K\in\mathcal{C}}
\ \sum_{k=1}^{K} s_{k, y_k}
\;-\;
\lambda \sum_{k=2}^{K} \mathbb{1}\!\left[y_k = y_{k-1}\right],
\]
with $\lambda\!\ge\!0$ fixed globally. Finally, $y_t=\hat y_k$ for all $t\in B_k$.

\noindent
All baselines share the same frozen encoder features and prompts, require no dataset–specific tuning, and are evaluated with the same metrics as our method. We find that $\lambda$ is best for all the datasets.

\begin{table}[htbp]
\centering
\renewcommand{\arraystretch}{1.2}
\setlength{\tabcolsep}{6pt}

\caption{\textbf{Ablation on Temporal Prior and L2 Norm:} Parentheses show drops (\textcolor{blue}{↓})) from Best. Results on GTEA Dataset.}
\label{tab:temporal_l2_ablation_inverted}

\begin{tabular}{l|ccc}
\toprule
Metric & Best & Temporal Prior (abl.) & L2 Norm (abl.) \\
\midrule
F1@10 & 21.33 & 1.12 {\scriptsize \textcolor{blue}{(↓20.21)}} & 2.36 {\scriptsize \textcolor{blue}{(↓18.97)}} \\
F1@25 & 13.76 & 0.56 {\scriptsize \textcolor{blue}{(↓13.20)}} & 1.12 {\scriptsize \textcolor{blue}{(↓12.64)}} \\
F1@50 & 5.44  & 0.19 {\scriptsize \textcolor{blue}{(↓5.25)}}  & 0.62 {\scriptsize \textcolor{blue}{(↓4.82)}} \\
Edit  & 51.07 & 5.04 {\scriptsize \textcolor{blue}{(↓46.03)}} & 7.23 {\scriptsize \textcolor{blue}{(↓43.84)}} \\
Acc   & 28.08 & 6.54 {\scriptsize \textcolor{blue}{(↓21.54)}} & 8.43 {\scriptsize \textcolor{blue}{(↓19.65)}} \\
Avg   & 23.14 & 2.69 {\scriptsize \textcolor{blue}{(↓20.45)}} & 3.95 {\scriptsize \textcolor{blue}{(↓19.19)}} \\
\bottomrule
\end{tabular}

\end{table}

\subsubsection{Implementation Details} We implement our method and baselines using PyTorch. For the canonical action ordering which is required by the optimal transport algorithm, we do a random action ordering - thus only needing action set information. For extracting the action label embeddings, we perform simplistic transformations of the labels such as "pour\_coffee" in breakfast dataset to "pour coffee". For GTEA dataset we attach verbs to construct short phrases using the annotation.  We select the hyperparameter values of optimal transport stage using the values given in \cite{xu2024temporally}. Specifically, we set $\epsilon = 0.07$, $\alpha = 0.5$, $r = 0.04$, 
$\lambda_{\text{frames}} = 0.11$, and $\lambda_{\text{actions}} = 0.01$. Different from \cite{xu2024temporally} which uses optimal transport in an unbalanced setting, we find in our grid-search that balanced formulation gives the strongest results.
We extract the per-frame Vision-Language Models (VLMs) features using an NVIDIA A6000 machine. We extract per-frame VLM features using off-the-shelf checkpoints without fine-tuning. Action embeddings are cached for efficiency. 
The chosen values are fixed for all experiments.
 We use 4 families of VLMs - Perception Encoders (PECore) \cite{bolya2025perception} SigLIP \cite{zhai2023sigmoid}, CLIP\cite{radford2021learning} and OpenCLIP \cite{ilharco2021openclip}. The details of the VLMs are mentioned in \cref{tab:vlm_sizes}.

\begin{table}[htbp]
\centering
\small
\setlength{\tabcolsep}{6pt}
\caption{ \textbf{VLM Details:} Parameter sizes (in millions) of the 14 VLM variants used in our experiments.}
\label{tab:vlm_sizes}
\begin{tabular}{l l c}
\toprule
Abbrev & Checkpoint & \#Params (M) \\
\midrule
SigLIP-M1 & so400m-p16-256-i18n & 877.96 \\
SigLIP-M2 & large-p16-384        & 652.48 \\
SigLIP-M3 & so400m-p14-384       & 1128.76 \\
\midrule
OpenCLIP-M1 & {\scriptsize ViT-B/32 (laion2B-s34B-b79K)} & 151.28 \\
OpenCLIP-M2 & {\scriptsize ViT-B/16 (laion2B-s34B-b88K)} & 149.62 \\
OpenCLIP-M3 & {\scriptsize ViT-L/14 (laion2B-s32B-b82K)} & 427.62 \\
OpenCLIP-M4 & {\scriptsize ViT-H/14 (laion2B-s32B-b79K)} & 986.11 \\
OpenCLIP-M5 & {\scriptsize ViT-g/14 (laion2B-s34B-b88K)} & 1366.68 \\
\midrule
CLIP-M1 & ViT-B/32 & 151.28 \\
CLIP-M2 & ViT-B/16 & 149.62 \\
CLIP-M3 & ViT-L/14 & 427.62 \\
\midrule
PECore-M1 & B/16-224  & 447.66 \\
PECore-M2 & L/14-336  & 671.14 \\
PECore-M3 & G/14-448  & 2419.27 \\
\bottomrule
\end{tabular}
\end{table}

\subsection{Performance Comparison}

\begin{table*}[th]
\centering
\renewcommand{\arraystretch}{1.15}
\setlength{\tabcolsep}{4pt} 

\caption{\textbf{Comparison of baselines and all VLM variants across three datasets} (50 Salads, GTEA, and Breakfast). Each Avg is the mean of F1@10, F1@25, F1@50, Edit, and Accuracy. Maximum average value for each dataset and each family is boldfaced.}
\label{tab:vlm_full_results_with_avg}

\begin{tabular}{l|cccc|cccc|cccc}
\toprule
\multirow{2}{*}{Method} 
& \multicolumn{4}{c|}{50 Salads} 
& \multicolumn{4}{c|}{GTEA} 
& \multicolumn{4}{c}{Breakfast} \\
 & F1@{10,25,50} & Edit & Acc & Avg 
 & F1@{10,25,50} & Edit & Acc & Avg
 & F1@{10,25,50} & Edit & Acc & Avg \\
\midrule
Random
 & 0.00 / 0.00 / 0.00 & 0.20 & 5.46 & 1.13
 & 0.12 / 0.01 / 0.01 & 2.06 & 5.61 & 1.56
 & 0.00 / 0.00 / 0.00 & 0.72 & 17.45 & 3.63 \\
ES\_mean 
 & 1.49 / 0.13 / 0.04 & 5.40 & 5.63 & 2.54
 & 7.08 / 3.99 / 1.67 & 13.00 & 5.99 & 6.35
 & 20.56 / 11.97 / 3.63 & 22.39 & 17.14 & 15.14 \\
ES\_vote 
 & 2.77 / 0.84 / 0.31 & 6.72 & 5.68 & 3.26
 & 7.32 / 4.63 / 1.87 & 12.68 & 5.96 & 6.49
 & 22.15 / 13.69 / 4.52 & 24.66 & 17.15 & 16.43 \\
ES\_nrp 
 & 6.81 / 5.90 / 2.73 & 8.00 & 7.24 & \textbf{6.14}
 & 7.75 / 5.35 / 2.01 & 8.29 & 12.96 & \textbf{7.27}
 & 24.38 / 19.46 / 8.73 & 36.35 & 11.83 & \textbf{20.15} \\
\midrule
Ours (SigLIP-M1) 
 & 42.6 / 31.4 / 14.1 & 88.7 & 31.5 & \textbf{41.7}
 & 21.1 / 13.0 / 4.9 & 51.2 & 28.3 & 23.7
 & 54.0 / 39.5 / 15.0 & 92.7 & 30.9 & 46.4 \\
Ours (SigLIP-M2) 
 & 42.4 / 31.3 / 14.2 & 87.9 & 31.4 & 41.4
 & 21.3 / 13.8 / 5.4 & 51.1 & 28.1 & \textbf{23.9}
 & 54.0 / 39.5 / 15.2 & 92.7 & 30.9 & \textbf{46.5} \\
Ours (SigLIP-M3) 
 & 42.7 / 30.5 / 14.3 & 87.3 & 31.3 & 41.2
 & 20.8 / 13.0 / 4.9 & 50.7 & 27.9 & 23.5
 & 53.8 / 39.2 / 15.1 & 92.5 & 30.8 & 46.3 \\
\midrule
Ours (OpenCLIP-M1)
 & 40.8 / 30.0 / 13.2 & 78.0 & 30.7 & 38.5
 & 15.2 / 9.6 / 3.6 & 34.0 & 20.4 & 16.6
 & 52.6 / 38.2 / 14.8 & 90.3 & 30.4 & 45.3 \\
Ours (OpenCLIP-M2)
 & 41.1 / 31.1 / 13.8 & 79.2 & 31.4 & \textbf{39.3}
 & 15.9 / 10.0 / 3.8 & 35.3 & 21.0 & \textbf{17.2}
 & 53.5 / 38.9 / 15.4 & 91.2 & 30.7 & \textbf{45.9} \\
Ours (OpenCLIP-M3)
 & 39.2 / 29.2 / 12.4 & 76.0 & 30.4 & 37.4
 & 14.5 / 9.3 / 3.4 & 32.7 & 20.4 & 16.1
 & 52.8 / 38.1 / 14.9 & 90.5 & 30.2 & 45.3 \\
Ours (OpenCLIP-M4)
 & 39.6 / 29.2 / 13.5 & 77.7 & 29.5 & 37.9
 & 14.8 / 9.4 / 3.5 & 33.1 & 20.2 & 16.2
 & 53.0 / 38.4 / 15.0 & 90.8 & 30.3 & 45.5 \\
Ours (OpenCLIP-M5)
 & 39.1 / 28.8 / 12.4 & 73.7 & 30.1 & 36.8
 & 14.4 / 9.1 / 3.3 & 31.9 & 19.9 & 15.7
 & 52.3 / 37.7 / 14.7 & 89.8 & 29.9 & 44.8 \\
\midrule
Ours (CLIP-M1)
 & 41.8 / 31.2 / 13.8 & 88.6 & 31.3 & 41.4
 & 19.7 / 13.0 / 4.9 & 44.7 & 25.6 & 21.6
 & 54.0 / 39.5 / 15.2 & 92.6 & 30.9 & 46.4 \\
Ours (CLIP-M2)
 & 42.6 / 30.9 / 14.5 & 88.0 & 31.5 & \textbf{41.5}
 & 20.1 / 13.3 / 5.1 & 45.2 & 25.9 & \textbf{21.9}
 & 54.1 / 39.6 / 15.3 & 92.8 & 31.0 & \textbf{46.6} \\
Ours (CLIP-M3)
 & 41.3 / 30.0 / 13.6 & 85.5 & 30.8 & 40.2
 & 19.2 / 12.7 / 4.7 & 43.9 & 25.2 & 21.2
 & 53.7 / 39.1 / 15.0 & 92.1 & 30.7 & 46.1 \\
\midrule
Ours (PECore-M1)
 & 39.8 / 28.4 / 13.6 & 79.1 & 30.3 & \textbf{38.2}
 & 15.3 / 10.3 / 3.5 & 37.0 & 22.5 & \textbf{17.7}
 & 53.9 / 39.4 / 15.4 & 91.6 & 30.7 & \textbf{46.2} \\
Ours (PECore-M2)
 & 38.8 / 26.7 / 11.6 & 68.8 & 28.5 & 34.9
 & 13.6 / 8.5 / 2.9 & 30.4 & 19.6 & 15.0
 & 52.1 / 37.3 / 14.5 & 88.7 & 29.6 & 44.4 \\
Ours (PECore-M3)
 & 39.3 / 28.5 / 13.3 & 75.2 & 29.3 & 37.1
 & 14.2 / 9.1 / 3.2 & 31.8 & 20.1 & 15.7
 & 53.0 / 38.5 / 15.1 & 89.9 & 30.1 & 45.3 \\
\bottomrule
\end{tabular}
\end{table*}

We report the baseline results and performance of our model in \cref{tab:vlm_full_results_with_avg}. Among the training-free baselines, ES\_NRP consistently achieves the strongest results across datasets, outperforming ES\_mean and ES\_vote. This confirms that equal-splits baselines benefit from a non-repetition constraint, validating the use of NRP as the most representative baseline.
GTEA consists of egocentric videos annotated with fine-grained action classes. The egocentric viewpoint introduces camera motion, making it the most challenging benchmark, which is reflected in the lowest performance. Across the five VLMs tested—SigLIP-M1, SigLIP-M2, OpenCLIP-M1, CLIP-M1, and PECore-M1—performance differences are relatively small on Breakfast and 50 Salads. In contrast, all models show larger drops on GTEA, reflecting the challenges posed by its egocentric viewpoint, rapid transitions, and large number of fine-grained classes. 
Our qualitative results are shown in \cref{fig:qualitative_bars}. 
Our results show that, our OVTAS pipeline significantly outperforms baselines, and establishes encouraging results for the novel task of Open-Vocabulary Zero-Shot Action Segmentation.

\subsection{Ablation Studies} To validate the components of our OVTAS pipeline, we perform ablation studies.

\begin{table*}[htbp]
\centering
\renewcommand{\arraystretch}{1.2}
\setlength{\tabcolsep}{6pt}

\caption{\textbf{Stage Ablation studies:} For each dataset, we compare the Best model with FAES (Random-ASOT) and SMTS (Baseline-0). Columns report F1 scores at different thresholds, Edit, Acc, and their mean (Avg). Drops from Best are shown in \textcolor{blue}{blue} with downward arrows (\textcolor{blue}{↓}).}
\label{tab:ablation_vertical}

\begin{tabular}{ll|cccccc}
\toprule
Dataset & Ablated Stage & F1@10 & F1@25 & F1@50 & Edit & Acc & Avg \\
\midrule
\multirow{3}{*}{50 Salads}
 & None & 41.84 & 31.19 & 13.84 & 88.58 & 31.30 & 41.81 \\
 & Stage1 ($\downarrow$) & 6.71 {\scriptsize \textcolor{blue}{(↓35.13)}} & 
     5.16 {\scriptsize \textcolor{blue}{(↓26.03)}} & 
     2.27 {\scriptsize \textcolor{blue}{(↓11.57)}} 
     & 9.35 {\scriptsize \textcolor{blue}{(↓79.23)}} & 
     28.53 {\scriptsize \textcolor{blue}{(↓2.77)}} & 
     10.40 {\scriptsize \textcolor{blue}{(↓31.41)}} \\
 & Stage2 ($\downarrow$) & 0.17 {\scriptsize \textcolor{blue}{(↓41.67)}} & 
     0.08 {\scriptsize \textcolor{blue}{(↓31.11)}} & 
     0.03 {\scriptsize \textcolor{blue}{(↓13.81)}} 
     & 0.88 {\scriptsize \textcolor{blue}{(↓87.70)}} & 
     5.48 {\scriptsize \textcolor{blue}{(↓25.82)}} & 
     3.01 {\scriptsize \textcolor{blue}{(↓38.80)}} \\
\midrule
\multirow{3}{*}{GTEA}
 & None & 21.33 & 13.76 & 5.44 & 51.07 & 28.08 & 23.14 \\
 & Stage1 ($\downarrow$) & 1.38 {\scriptsize \textcolor{blue}{(↓19.95)}} & 
     0.60 {\scriptsize \textcolor{blue}{(↓13.16)}} & 
     0.25 {\scriptsize \textcolor{blue}{(↓5.19)}} 
     & 17.39 {\scriptsize \textcolor{blue}{(↓33.68)}} & 
     17.30 {\scriptsize \textcolor{blue}{(↓10.78)}} & 
     7.38 {\scriptsize \textcolor{blue}{(↓15.76)}} \\
 & Stage2 ($\downarrow$) & 1.18 {\scriptsize \textcolor{blue}{(↓20.15)}} & 
     0.52 {\scriptsize \textcolor{blue}{(↓13.24)}} & 
     0.16 {\scriptsize \textcolor{blue}{(↓5.28)}} 
     & 3.95 {\scriptsize \textcolor{blue}{(↓47.12)}} & 
     5.41 {\scriptsize \textcolor{blue}{(↓22.67)}} & 
     2.64 {\scriptsize \textcolor{blue}{(↓20.50)}} \\
\midrule
\multirow{3}{*}{Breakfast}
 & None & 54.26 & 39.60 & 15.34 & 92.44 & 30.89 & 46.51 \\
 & Stage1 ($\downarrow$) & 12.58 {\scriptsize \textcolor{blue}{(↓41.68)}} & 
     8.36 {\scriptsize \textcolor{blue}{(↓31.24)}} & 
     2.85 {\scriptsize \textcolor{blue}{(↓12.49)}} 
     & 18.87 {\scriptsize \textcolor{blue}{(↓73.57)}} & 
     26.08 {\scriptsize \textcolor{blue}{(↓4.81)}} & 
     13.75 {\scriptsize \textcolor{blue}{(↓32.76)}} \\
 & Stage2 ($\downarrow$) & 1.16 {\scriptsize \textcolor{blue}{(↓53.10)}} & 
     0.67 {\scriptsize \textcolor{blue}{(↓38.93)}} & 
     0.27 {\scriptsize \textcolor{blue}{(↓15.07)}} 
     & 8.13 {\scriptsize \textcolor{blue}{(↓84.31)}} & 
     17.50 {\scriptsize \textcolor{blue}{(↓13.39)}} & 
     5.95 {\scriptsize \textcolor{blue}{(↓40.56)}} \\
\bottomrule
\end{tabular}

\end{table*}

\subsubsection{Stage Ablation} We ablate the Stage 1, by randomly permuting the frame features and action embedding features, and run the pipeline and measure the drop in performance for both the stages. For ablating stage 2, we perform frame level predictions using maximum probability label for for each frame's action classification probabilities. Our results show significant drops for all the metrics and for all the datasets - indicating the criticality of both the stages towards final performance.
\subsubsection{L2-Norm and Temporal Prior Ablation} We also independently ablate L2-norm in stage1, and use of temporal priors in stage 2. We present our results on GTEA dataset. \cref{tab:temporal_l2_ablation_inverted}. Our results demonstrate the criticality of both those design choices for strong performance.

\begin{table}[t]
\centering
\renewcommand{\arraystretch}{1.15}
\setlength{\tabcolsep}{5pt}
\caption{ \textbf{Performance variation by video duration:} across different video length bins (in seconds) for Breakfast, GTEA, and 50 Salads datasets.}
\label{tab:bin_length_perf}
\begin{tabular}{l|c c c c c c}
\toprule
\textbf{Breakfast} & Acc & Edit & F1@10 & F1@25 & F1@50 & Avg \\
\midrule
0--60s   & 46.31 & 98.51 & 76.05 & 62.39 & 29.16 & 62.48 \\
60--120s & 40.99 & 96.28 & 63.98 & 50.89 & 20.26 & 54.48 \\
$\geq$120s & 26.41 & 82.37 & 41.40 & 25.81 & 8.16  & 36.83 \\
\midrule
\textbf{GTEA} & Acc & Edit & F1@10 & F1@25 & F1@50 & Avg \\
\midrule
0--60s   & 27.75 & 50.65 & 24.38 & 17.14 & 8.38  & 25.66 \\
60--120s & 23.31 & 33.19 & 13.54 & 8.12  & 3.05  & 16.24 \\
$\geq$120s & 16.82 & 12.99 & 6.25  & 2.27  & 0.57  & 7.78  \\
\midrule
\textbf{50 Salads} & Acc & Edit & F1@10 & F1@25 & F1@50 & Avg \\
\midrule
240--360s & 30.57 & 84.37 & 42.15 & 31.39 & 13.73 & 40.44 \\
360--480s & 32.92 & 75.01 & 41.73 & 31.34 & 14.96 & 39.19 \\
$\geq$480s & 23.40 & 76.11 & 28.31 & 19.88 & 7.41  & 31.02 \\
\bottomrule
\end{tabular}
\end{table}

\begin{figure*}[t]
\centering
\setlength{\tabcolsep}{4pt} 
\renewcommand{\arraystretch}{1.0}
\begin{tabular}{ccc}

\includegraphics[width=0.30\textwidth]{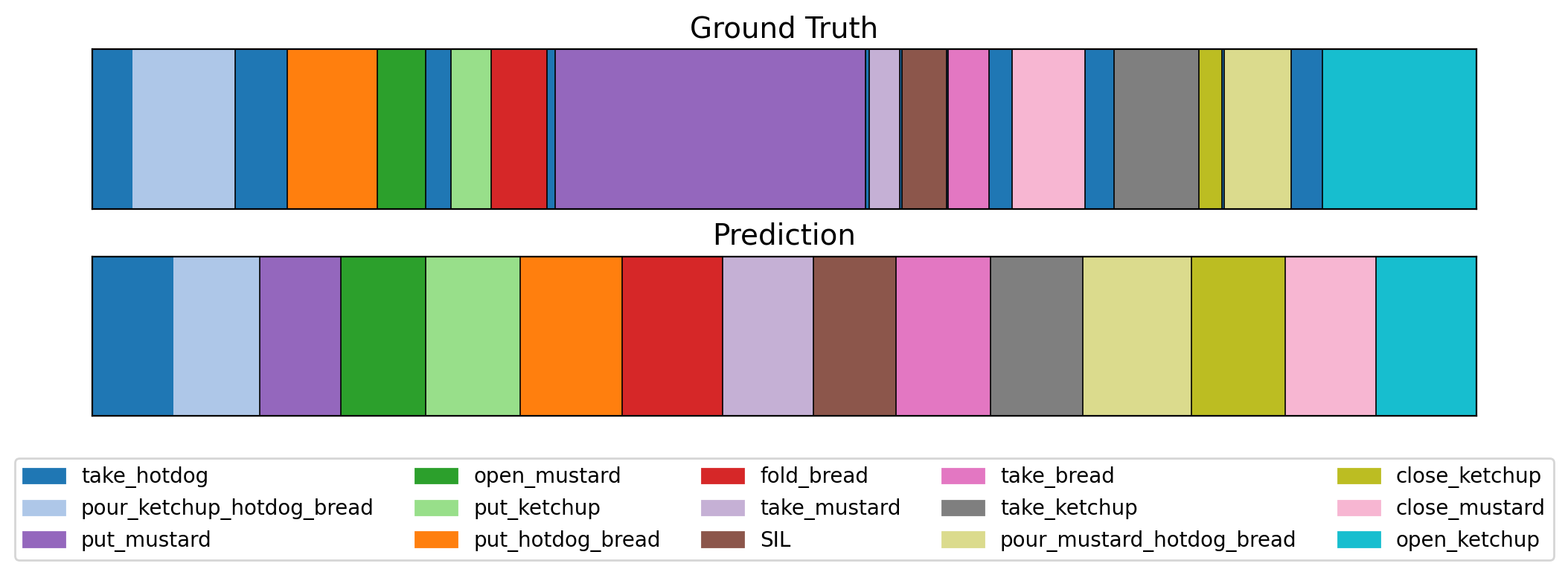} &
\includegraphics[width=0.30\textwidth]{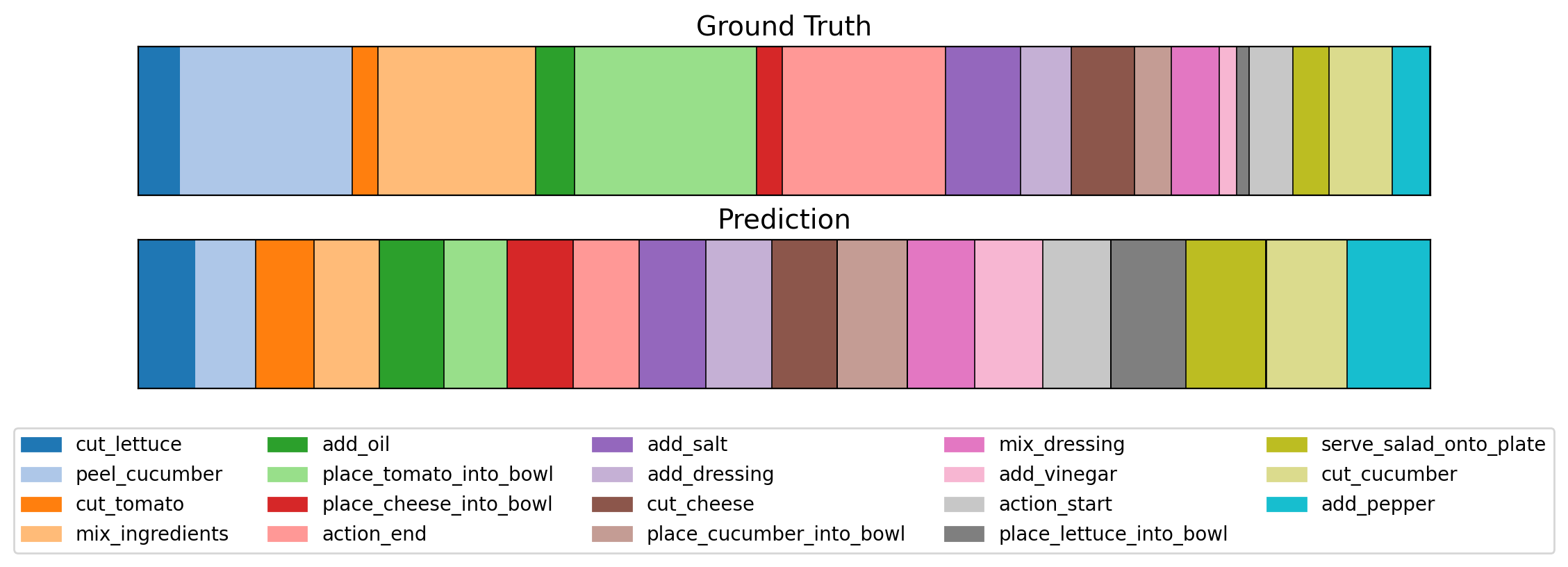} &
\includegraphics[width=0.30\textwidth]{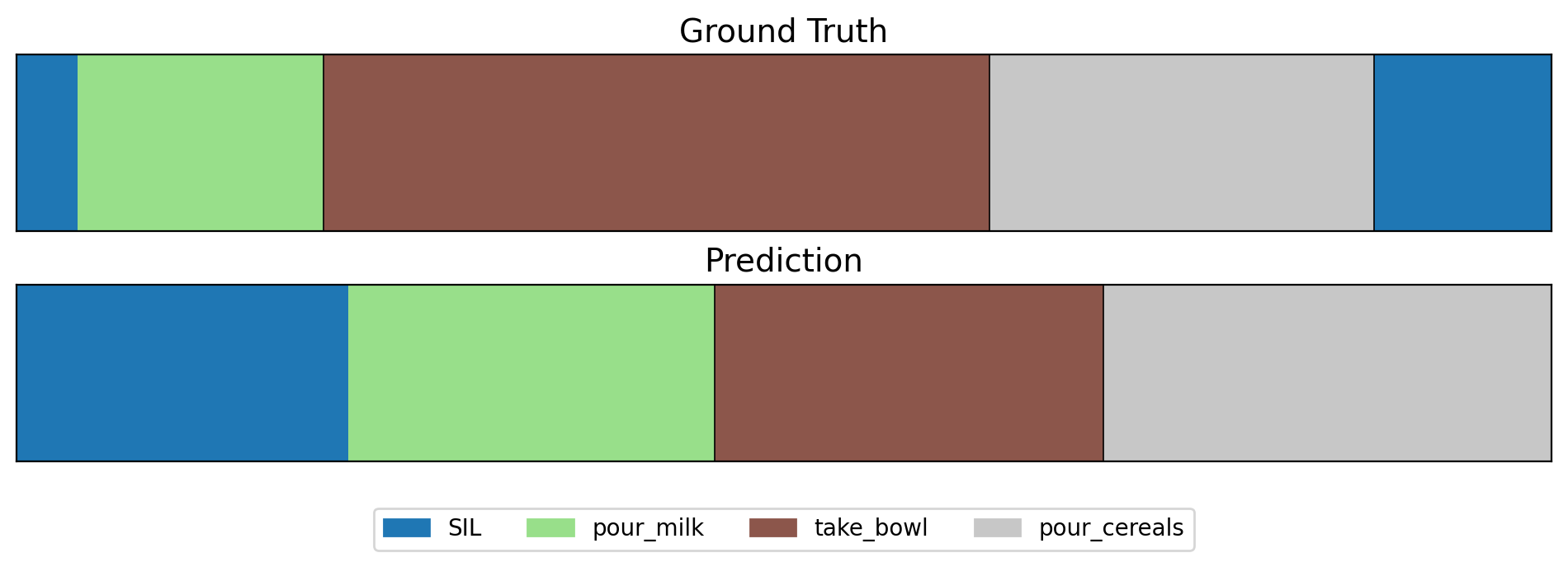} \\[4pt]

\includegraphics[width=0.30\textwidth]{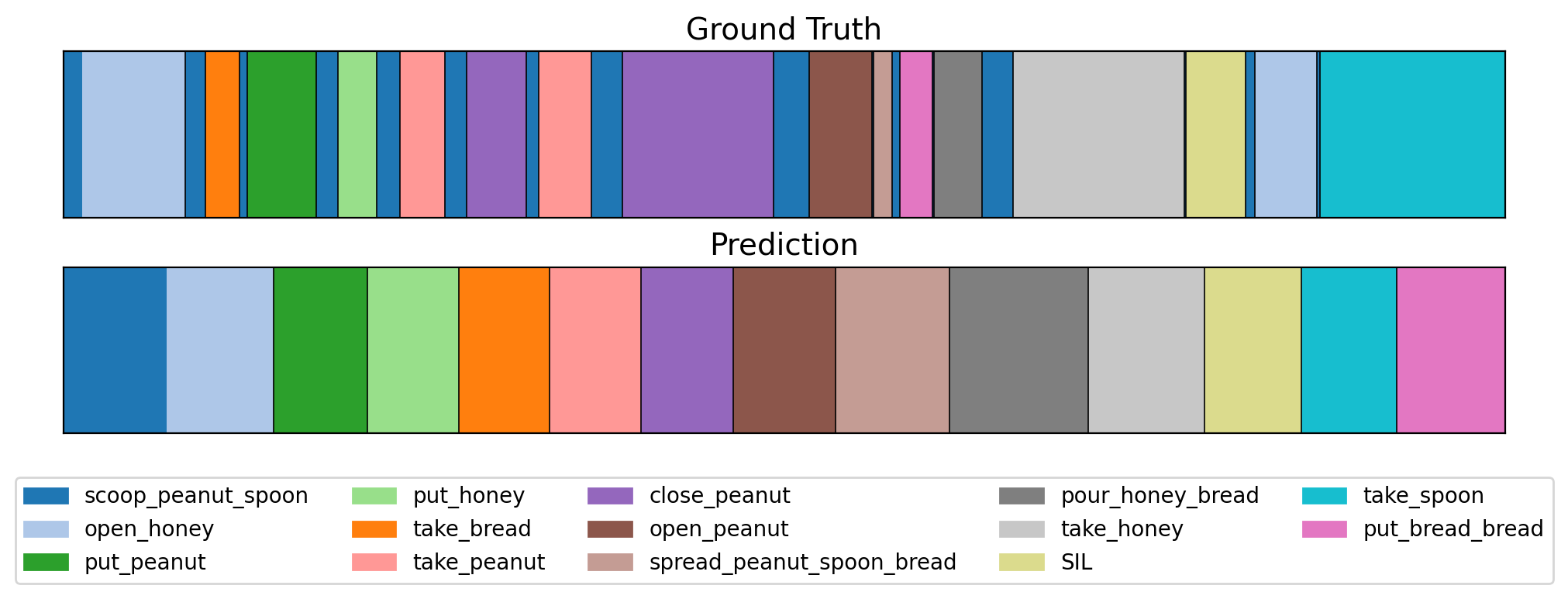} &
\includegraphics[width=0.30\textwidth]{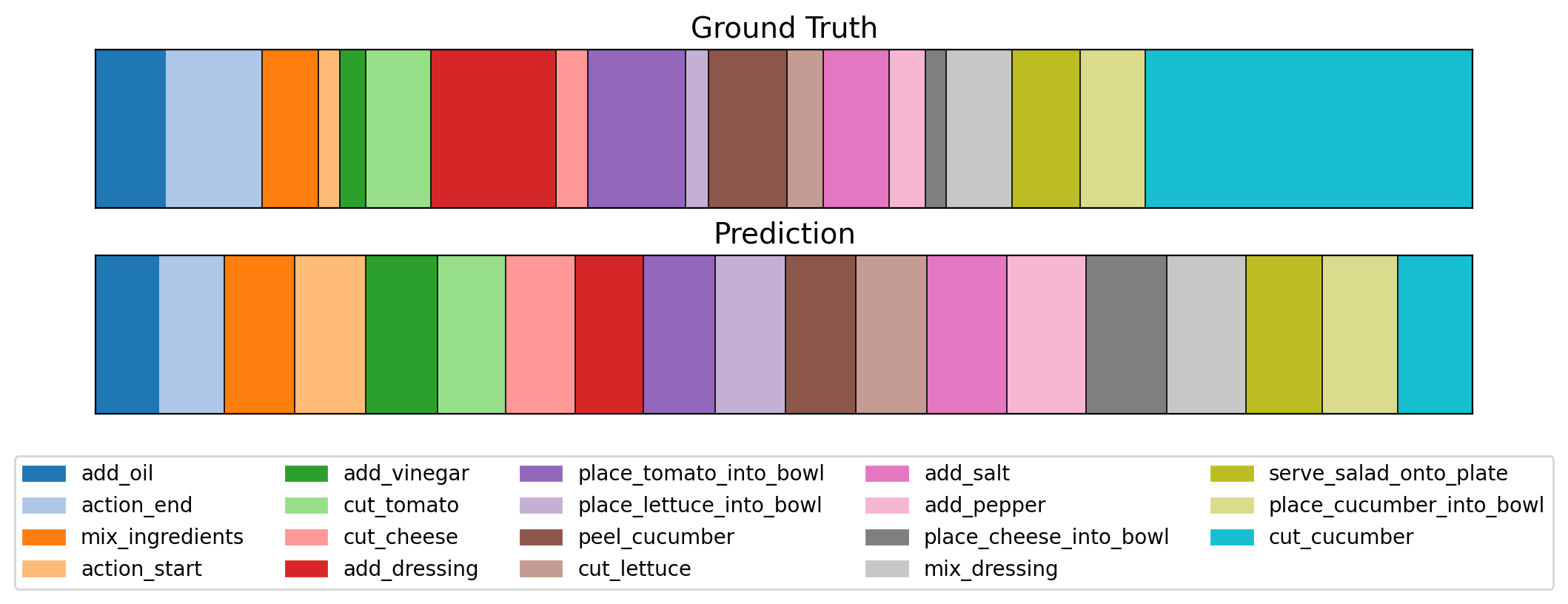} &
\includegraphics[width=0.30\textwidth]{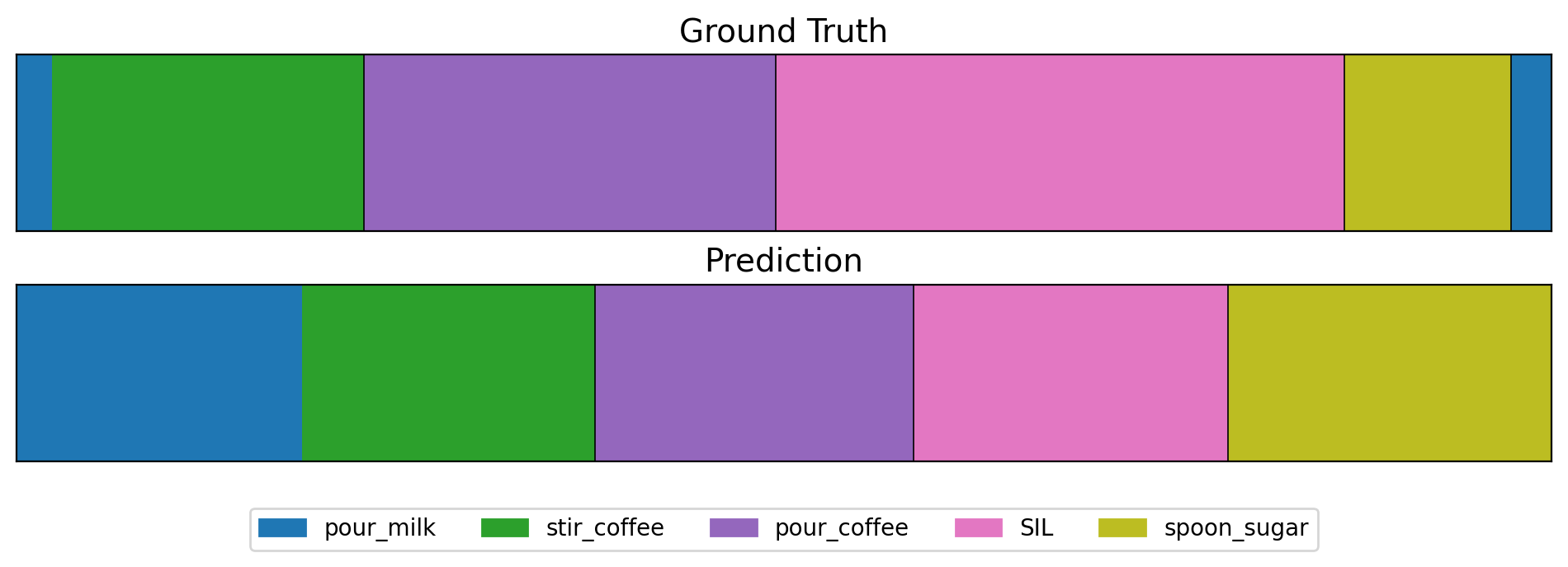} \\
\end{tabular}
\caption{\textbf{Qualitative results:} columns show segmentation results of our method on GTEA, 50 Salads, and Breakfast (left to right), with two examples each.}

\label{fig:qualitative_bars}
\end{figure*}

\begin{table}[t]
\centering
\renewcommand{\arraystretch}{1.15}
\setlength{\tabcolsep}{6pt}
\caption{\textbf{Performance variation by number of action segments in ground truth:} Performance across bins of ground-truth segment counts for GTEA, Breakfast, and 50 Salads datasets. }
\label{tab:segcount_perf}
\begin{tabular}{l|c c c c c c}
\toprule
\textbf{GTEA} & Acc & Edit & F1@10 & F1@25 & F1@50 & Avg \\
\midrule
20--29 & 28.23 & 72.77 & 48.55 & 33.53 & 16.18 & 39.85 \\
30--39 & 23.12 & 42.79 & 19.11 & 11.78 & 4.78  & 20.32 \\
40--49 & 22.19 & 15.14 & 10.06 & 5.95  & 2.27  & 11.12 \\
\midrule
\textbf{Breakfast} & Acc & Edit & F1@10 & F1@25 & F1@50 & Avg \\
\midrule
0--4   & 42.79 & 98.69 & 71.90 & 59.28 & 26.25 & 59.78 \\
5--9   & 29.00 & 88.56 & 50.52 & 35.06 & 12.60 & 43.15 \\
10--14 & 25.11 & 72.61 & 38.81 & 23.16 & 7.82  & 33.50 \\
15--19 & 22.14 & 32.30 & 21.54 & 12.73 & 4.35  & 18.61 \\
\midrule
\textbf{50 Salads} & Acc & Edit & F1@10 & F1@25 & F1@50 & Avg \\
\midrule
15--19 & 25.73 & 83.77 & 32.79 & 23.86 & 10.06 & 35.24 \\
20--24 & 32.74 & 77.48 & 44.37 & 33.12 & 14.08 & 40.36 \\
25--29 & 32.00 & 65.57 & 38.57 & 24.62 & 14.18 & 34.99 \\
\bottomrule
\end{tabular}
\end{table}



\section{Discussions}

\subsubsection{VLM analysis} 
\begin{figure}[htbp]
\centering
\includegraphics[width=1.05\linewidth]{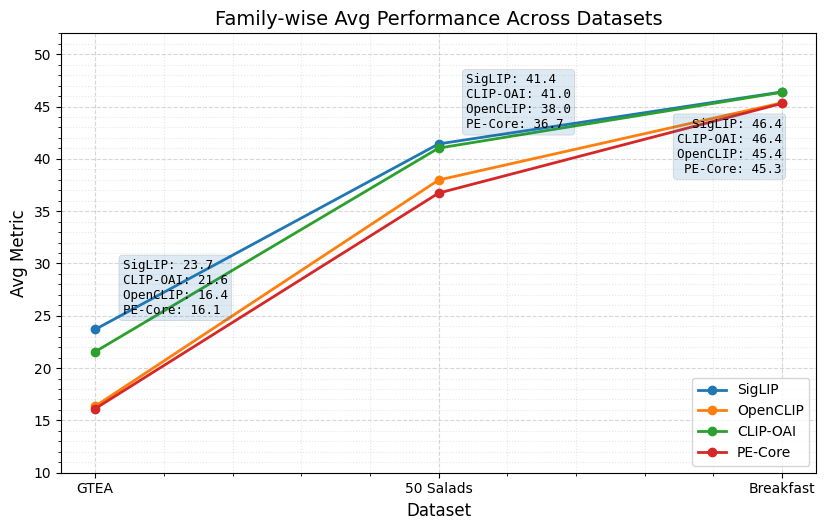}
\caption{ \textbf{VLM Family average of Avg metric:} across datasets (GTEA, 50 Salads, Breakfast). Each line is a VLM family; per-dataset boxes list all values.}
\label{fig:family-avg-lineplot}
\end{figure}

In this section we analyze the impact of VLM family choice and model size on OVTAS performance. We first compare different families of VLMs to understand which architectures and pre-training strategies 
are most effective for temporal action segmentation. 
We then study the effect of scaling model size within each family to examine whether larger checkpoints bring consistent improvements.
\paragraph{VLM Family Analysis}
We compare the four VLM families (SigLIP, CLIP, OpenCLIP, and PECore) by averaging their performance across the three benchmark datasets. 
As shown in \cref{fig:family-avg-lineplot}, consistent trends are observed across all datasets with the same ranking: 
the SigLIP family outperforms all others, followed by CLIP, while OpenCLIP and PECore trail behind. 
This consistency indicates that the relative strength of each family is stable across domains, 
with SigLIP providing the most reliable performance for OVTAS.

\paragraph{VLM Size Analysis}

\begin{figure}[htbp]
    \centering
    \includegraphics[width=1.00\linewidth]{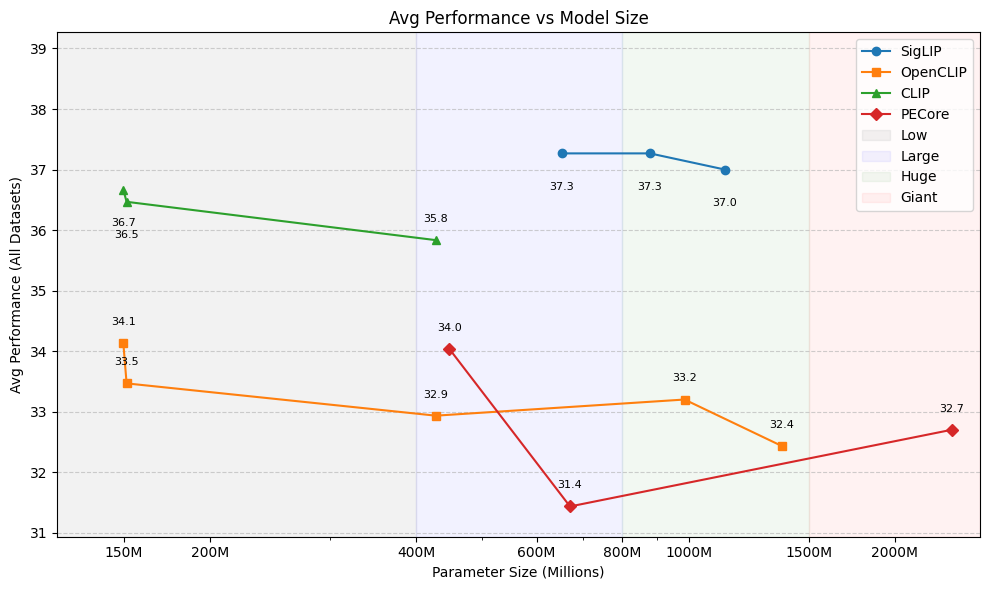}
    \caption{\textbf{Performance vs Model Size.} 
    Models are grouped by family (SigLIP, CLIP, OpenCLIP, PECore). 
    Shaded regions indicate parameter size bins: Low ($\leq$400M), 
    Large (400--800M), Huge (800--1500M), and Giant ($>$1500M).}
    \label{fig:global_avg_vs_params}
\end{figure}

The plot in \cref{fig:global_avg_vs_params} shows that simply scaling VLMs up does not yield better action-segmentation performance using OVTAS—in all VLM families. Larger checkpoints underperform their smaller counterparts. Thus, as future research direction, we can take up (i) stronger text prompting and (ii) video frame pre-processing such as cropping. Together these may extract frame and action label embeddings possibly giving performance increase with increasing VLM size.

\subsubsection{Effect of Video Length and Action Segment Counts}
\paragraph{}
We analyze the effect of video duration on performance by binning videos into different length ranges. Results in Table~\ref{tab:bin_length_perf} show a consistent trend across datasets: as videos become longer, the performance drops. For example, in Breakfast the highest accuracy and F1 scores are observed for shorter clips ($<60$s), while performance steadily degrades for videos exceeding $120$s. A similar pattern is found in GTEA and 50 Salads, suggesting that longer videos introduce greater temporal variability and error propagation in training-free segmentation. We also study the effect of the number of ground-truth action segments per video (Table~\ref{tab:segcount_perf}). For GTEA, where each video contains many short segments (mean $\sim$36), the model struggles compared to Breakfast, where the mean is $\sim$5 segments. 50 Salads lies between these two extremes. This shows that the number of fine-grained action boundaries strongly influences performance, with dense sequences of short actions being particularly challenging.

\begin{table}[htbp]
\centering
\renewcommand{\arraystretch}{1.15}
\setlength{\tabcolsep}{5pt}
\caption{\textbf{Statistics of segment durations} across datasets.}
\label{tab:segment_durations}
\begin{tabular}{l|c c c}
\toprule
\textbf{Dataset} & Min (s) & Max (s) & Mean (s) \\
\midrule
GTEA      & 0.07  & 44.73  & 1.94  \\
Breakfast & 0.07  & 386.00 & 20.95 \\
50 Salads & 0.03  & 138.47 & 18.59 \\
\bottomrule
\end{tabular}
\end{table}

\paragraph{}
Finally, Table~\ref{tab:segment_durations} summarizes segment duration statistics. The mean segment duration in GTEA is only 1.94s, compared to 20.95s in Breakfast and 18.59s in 50 Salads. These results explain the difficulty in GTEA: the model must repeatedly localize boundaries within very short spans, leaving little room for temporal context aggregation. In contrast, datasets with longer average segments allow the model to achieve better segmentation alignment.




\subsubsection{Limitations and Future Work} While we perform our zero-shot open-vocabulary action segmentation on 3 prominent action segmentation datasets, future work can explore action segmentation datasets. Further, future works can dive deeper into prompt engineering in order to improve the performance. Also enhancing the temporal modeling ability of the optimal transport algorithm is important for such training-free action segmentation pipelines.

\section{Conclusion}
We presented \textit{OVTAS}, a pipeline for temporal action segmentation that is simultaneously open-vocabulary, zero-shot, and training-free. By leveraging the strong zero-shot capabilities of vision–language models, OVTAS segments videos into actions without requiring any task-specific supervision. Our two-stage design—Frame–Action Embedding Similarity (FAES) followed by Similarity-Matrix Temporal Segmentation (SMTS)—delivers encouraging action segmentation performance across standard TAS benchmarks. Beyond its empirical results, OVTAS highlights the promise of adapting pretrained VLMs for structured temporal understanding, enabling action segmentation in an open-vocabulary setting. We also release the extracted embeddings of 14 VLMs across three datasets, providing ready-to-use features that lower computational barriers and enable further research into open-vocabulary and zero-shot action segmentation. We believe this work opens new directions for applying VLMs for action segmentation.


{
    \small
    \bibliographystyle{ieeetr}
    \bibliography{references}

@inproceedings{xu2024temporally,
  title={Temporally consistent unbalanced optimal transport for unsupervised action segmentation},
  author={Xu, Ming and Gould, Stephen},
  booktitle={Proceedings of the IEEE/CVF Conference on Computer Vision and Pattern Recognition},
  pages={14618--14627},
  year={2024}
}

@article{ding2023temporal,
  title={Temporal action segmentation: An analysis of modern techniques},
  author={Ding, Guodong and Sener, Fadime and Yao, Angela},
  journal={IEEE Transactions on Pattern Analysis and Machine Intelligence},
  volume={46},
  number={2},
  pages={1011--1030},
  year={2023},
  publisher={IEEE}
}

@inproceedings{li2021temporal,
  title={Temporal action segmentation from timestamp supervision},
  author={Li, Zhe and Abu Farha, Yazan and Gall, Jurgen},
  booktitle={Proceedings of the IEEE/CVF Conference on Computer Vision and Pattern Recognition},
  pages={8365--8374},
  year={2021}
}

@inproceedings{khan2022timestamp,
  title={Timestamp-supervised action segmentation with graph convolutional networks},
  author={Khan, Hamza and Haresh, Sanjay and Ahmed, Awais and Siddiqui, Shakeeb and Konin, Andrey and Zia, M Zeeshan and Tran, Quoc-Huy},
  booktitle={2022 IEEE/RSJ International Conference on Intelligent Robots and Systems (IROS)},
  pages={10619--10626},
  year={2022},
  organization={IEEE}
}

@inproceedings{richard2018action,
  title={Action sets: Weakly supervised action segmentation without ordering constraints},
  author={Richard, Alexander and Kuehne, Hilde and Gall, Juergen},
  booktitle={Proceedings of the IEEE conference on Computer Vision and Pattern Recognition},
  pages={5987--5996},
  year={2018}
}

@inproceedings{li2020set,
  title={Set-constrained viterbi for set-supervised action segmentation},
  author={Li, Jun and Todorovic, Sinisa},
  booktitle={Proceedings of the IEEE/CVF Conference on Computer Vision and Pattern Recognition},
  pages={10820--10829},
  year={2020}
}

@inproceedings{richard2017weakly,
  title={Weakly supervised action learning with rnn based fine-to-coarse modeling},
  author={Richard, Alexander and Kuehne, Hilde and Gall, Juergen},
  booktitle={Proceedings of the IEEE conference on Computer Vision and Pattern Recognition},
  pages={754--763},
  year={2017}
}

@article{kuehne2017weakly,
  title={Weakly supervised learning of actions from transcripts},
  author={Kuehne, Hilde and Richard, Alexander and Gall, Juergen},
  journal={Computer Vision and Image Understanding},
  volume={163},
  pages={78--89},
  year={2017},
  publisher={Elsevier}
}

@inproceedings{ding2022leveraging,
  title={Leveraging action affinity and continuity for semi-supervised temporal action segmentation},
  author={Ding, Guodong and Yao, Angela},
  booktitle={European Conference on Computer Vision},
  pages={17--32},
  year={2022},
  organization={Springer}
}

@inproceedings{radford2021learning,
  title={Learning transferable visual models from natural language supervision},
  author={Radford, Alec and Kim, Jong Wook and Hallacy, Chris and Ramesh, Aditya and Goh, Gabriel and Agarwal, Sandhini and Sastry, Girish and Askell, Amanda and Mishkin, Pamela and Clark, Jack and others},
  booktitle={International conference on machine learning},
  pages={8748--8763},
  year={2021},
  organization={PmLR}
}

@inproceedings{zhai2023sigmoid,
  title={Sigmoid loss for language image pre-training},
  author={Zhai, Xiaohua and Mustafa, Basil and Kolesnikov, Alexander and Beyer, Lucas},
  booktitle={Proceedings of the IEEE/CVF international conference on computer vision},
  pages={11975--11986},
  year={2023}
}

@article{bolya2025perception,
  title={Perception encoder: The best visual embeddings are not at the output of the network},
  author={Bolya, Daniel and Huang, Po-Yao and Sun, Peize and Cho, Jang Hyun and Madotto, Andrea and Wei, Chen and Ma, Tengyu and Zhi, Jiale and Rajasegaran, Jathushan and Rasheed, Hanoona and others},
  journal={arXiv preprint arXiv:2504.13181},
  year={2025}
}

@inproceedings{fathi2011learning,
  title={Learning to recognize objects in egocentric activities},
  author={Fathi, Alireza and Ren, Xiaofeng and Rehg, James M},
  booktitle={Proceedings of the IEEE Conference on Computer Vision and Pattern Recognition (CVPR)},
  pages={3281--3288},
  year={2011}
}

@inproceedings{stein2013combining,
  title={Combining embedded accelerometers with computer vision for recognizing food preparation activities},
  author={Stein, Sebastian and McKenna, Stephen J},
  booktitle={Proceedings of the ACM International Joint Conference on Pervasive and Ubiquitous Computing (UbiComp)},
  pages={729--738},
  year={2013}
}

@inproceedings{kuehne2014language,
  title={The language of actions: Recovering the syntax and semantics of goal-directed human activities},
  author={Kuehne, Hilde and Arslan, Ali and Serre, Thomas},
  booktitle={Proceedings of the IEEE Conference on Computer Vision and Pattern Recognition (CVPR)},
  pages={780--787},
  year={2014}
}

@inproceedings{li2021action,
  title={Action shuffle alternating learning for unsupervised action segmentation},
  author={Li, Jun and Todorovic, Sinisa},
  booktitle={Proceedings of the IEEE/CVF Conference on Computer Vision and Pattern Recognition},
  pages={12628--12636},
  year={2021}
}

@inproceedings{richard2018neuralnetwork,
  title={Neuralnetwork-viterbi: A framework for weakly supervised video learning},
  author={Richard, Alexander and Kuehne, Hilde and Iqbal, Ahsan and Gall, Juergen},
  booktitle={Proceedings of the IEEE conference on Computer Vision and Pattern Recognition},
  pages={7386--7395},
  year={2018}
}

@inproceedings{swetha2021unsupervised,
  title={Unsupervised discriminative embedding for sub-action learning in complex activities},
  author={Swetha, Sirnam and Kuehne, Hilde and Rawat, Yogesh S and Shah, Mubarak},
  booktitle={2021 IEEE International Conference on Image Processing (ICIP)},
  pages={2588--2592},
  year={2021},
  organization={IEEE}
}

@inproceedings{ishikawa2021alleviating,
  title={Alleviating over-segmentation errors by detecting action boundaries},
  author={Ishikawa, Yuchi and Kasai, Seito and Aoki, Yoshimitsu and Kataoka, Hirokatsu},
  booktitle={Proceedings of the IEEE/CVF winter conference on applications of computer vision},
  pages={2322--2331},
  year={2021}
}

@inproceedings{wang2020boundary,
  title={Boundary-aware cascade networks for temporal action segmentation},
  author={Wang, Zhenzhi and Gao, Ziteng and Wang, Limin and Li, Zhifeng and Wu, Gangshan},
  booktitle={European Conference on Computer Vision},
  pages={34--51},
  year={2020},
  organization={Springer}
}

@inproceedings{spurio2025hierarchical,
  title={Hierarchical vector quantization for unsupervised action segmentation},
  author={Spurio, Federico and Bahrami, Emad and Francesca, Gianpiero and Gall, Juergen},
  booktitle={Proceedings of the AAAI Conference on Artificial Intelligence},
  volume={39},
  number={7},
  pages={6996--7005},
  year={2025}
}

@article{ilharco2021openclip,
  title={Openclip},
  author={Ilharco, Gabriel and Wortsman, Mitchell and Carlini, Nicholas and Taori, Rohan and Dave, Achal and Shankar, Vaishaal and Namkoong, Hongseok and Miller, John and Hajishirzi, Hannaneh and Farhadi, Ali and others},
  journal={Zenodo},
  year={2021}
}

@inproceedings{sarfraz2021temporally,
  title={Temporally-weighted hierarchical clustering for unsupervised action segmentation},
  author={Sarfraz, Saquib and Murray, Naila and Sharma, Vivek and Diba, Ali and Van Gool, Luc and Stiefelhagen, Rainer},
  booktitle={Proceedings of the IEEE/CVF Conference on Computer Vision and Pattern Recognition},
  pages={11225--11234},
  year={2021}
}

@article{romeo2025multi,
  title={Multi-modal temporal action segmentation for manufacturing scenarios},
  author={Romeo, Laura and Marani, Roberto and Perri, Anna Gina and Gall, Juergen},
  journal={Engineering Applications of Artificial Intelligence},
  volume={148},
  pages={110320},
  year={2025},
  publisher={Elsevier}
}

@article{de2021first,
  title={A first evaluation of a multi-modal learning system to control surgical assistant robots via action segmentation},
  author={De Rossi, Giacomo and Minelli, Marco and Roin, Serena and Falezza, Fabio and Sozzi, Alessio and Ferraguti, Federica and Setti, Francesco and Bonf{\`e}, Marcello and Secchi, Cristian and Muradore, Riccardo},
  journal={IEEE Transactions on Medical Robotics and Bionics},
  volume={3},
  number={3},
  pages={714--724},
  year={2021},
  publisher={IEEE}
}

@article{chen2025backbone,
  title={A backbone for long-horizon robot task understanding},
  author={Chen, Xiaoshuai and Chen, Wei and Lee, Dongmyoung and Ge, Yukun and Rojas, Nicolas and Kormushev, Petar},
  journal={IEEE Robotics and Automation Letters},
  year={2025},
  publisher={IEEE}
}

@inproceedings{okamoto2024hierarchical,
  title={Hierarchical neurosymbolic approach for comprehensive and explainable action quality assessment},
  author={Okamoto, Lauren and Parmar, Paritosh},
  booktitle={Proceedings of the IEEE/CVF Conference on computer vision and pattern recognition},
  pages={3204--3213},
  year={2024}
}

@inproceedings{bueno2023leveraging,
  title={Leveraging triplet loss for unsupervised action segmentation},
  author={Bueno-Benito, Elena and Vecino, Biel Tura and Dimiccoli, Mariella},
  booktitle={Proceedings of the IEEE/CVF Conference on Computer Vision and Pattern Recognition},
  pages={4922--4930},
  year={2023}
}

@inproceedings{wang2022sscap,
  title={Sscap: Self-supervised co-occurrence action parsing for unsupervised temporal action segmentation},
  author={Wang, Zhe and Chen, Hao and Li, Xinyu and Liu, Chunhui and Xiong, Yuanjun and Tighe, Joseph and Fowlkes, Charless},
  booktitle={Proceedings of the IEEE/CVF Winter Conference on Applications of Computer Vision},
  pages={1819--1828},
  year={2022}
}

@article{gammulle2020fine,
  title={Fine-grained action segmentation using the semi-supervised action GAN},
  author={Gammulle, Harshala and Denman, Simon and Sridharan, Sridha and Fookes, Clinton},
  journal={Pattern Recognition},
  volume={98},
  pages={107039},
  year={2020},
  publisher={Elsevier}
}

@article{song2025unsupervised,
  title={Unsupervised Action Segmentation via Multi-scale Temporal-interaction Enhancement},
  author={Song, Zhiying and Chen, Kaixuan and Wang, Pengfei and Song, Mingli and Zheng, Nenggan},
  journal={IEEE Transactions on Circuits and Systems for Video Technology},
  year={2025},
  publisher={IEEE}
}

@article{knight2008sinkhorn,
  title={The Sinkhorn--Knopp algorithm: convergence and applications},
  author={Knight, Philip A},
  journal={SIAM Journal on Matrix Analysis and Applications},
  volume={30},
  number={1},
  pages={261--275},
  year={2008},
  publisher={SIAM}
}

@inproceedings{ju2022prompting,
  title={Prompting visual-language models for efficient video understanding},
  author={Ju, Chen and Han, Tengda and Zheng, Kunhao and Zhang, Ya and Xie, Weidi},
  booktitle={European conference on computer vision},
  pages={105--124},
  year={2022},
  organization={Springer}
}

@inproceedings{yu2025building,
  title={Building a multi-modal spatiotemporal expert for zero-shot action recognition with clip},
  author={Yu, Yating and Cao, Congqi and Zhang, Yueran and Lv, Qinyi and Min, Lingtong and Zhang, Yanning},
  booktitle={Proceedings of the AAAI Conference on Artificial Intelligence},
  volume={39},
  number={9},
  pages={9689--9697},
  year={2025}
}
}

\end{document}